\documentclass[conference]{IEEEtran}
\IEEEoverridecommandlockouts

\usepackage{cite}
\usepackage{amsmath,amssymb,amsfonts}
\usepackage{algorithmic}
\usepackage{graphicx}
\usepackage{textcomp}
\usepackage{xcolor}
\usepackage{url}
\def\BibTeX{{\rm B\kern-.05em{\sc i\kern-.025em b}\kern-.08em
    T\kern-.1667em\lower.7ex\hbox{E}\kern-.125emX}}

\begin{document}

\title{When Does Persona Prompting Actually Help? A Retrieval and Metric Analysis of Expert Role Injection in LLMs}

\author{ \IEEEauthorblockN{ Shuai Xiao\IEEEauthorrefmark{1}, Su Liu\IEEEauthorrefmark{1}, Weikai Zhou\IEEEauthorrefmark{1}, Jialun Wu\IEEEauthorrefmark{1}, Xinjie He\IEEEauthorrefmark{1}, Zhiyuan Lin\IEEEauthorrefmark{1}, Qiyang Xie\IEEEauthorrefmark{1} } \IEEEauthorblockA{ \IEEEauthorrefmark{1}Independent Researchers } }

\maketitle

\bstctlcite{BSTcontrol}

\begin{abstract}
Persona prompting is widely used to steer large language models, yet its practical value remains unclear. Prior work often evaluates persona prompting using aggregate scores, making it difficult to determine whether expert-role prompting consistently improves response quality or instead changes responses along different quality dimensions. We study this question through a controlled comparison of four prompting conditions across 1,140 open-ended questions spanning 38 expert roles and six domains: no role prompt, a generic domain-expert prompt, embedding-based role retrieval, and a hybrid retrieval method combining embedding search with LLM-based role selection.

Aggregate results show only small overall differences between conditions. However, metric-level analysis reveals a consistent tradeoff that aggregate averages obscure: role prompting systematically increases expertise depth while reducing clarity. These effects are highly conditional rather than universal. Role prompting performs best on advisory questions and in domains such as medicine and psychology, where structured expert framing and risk communication are intrinsically valuable. In contrast, baseline prompting performs better on conceptual and explanatory questions in finance, legal, science, and technology domains, where concise plain-language explanation is more important.

We further show that hybrid retrieval significantly improves over embedding-only role selection, although better role retrieval does not eliminate the broader expertise-depth versus clarity tradeoff. Overall, our findings suggest that persona prompting primarily reshapes response characteristics rather than broadly improving capability, and that multi-metric evaluation is necessary for understanding its effects.
\end{abstract}


\section{Introduction}
Large language models (LLMs) are increasingly used through prompting rather than task-specific fine-tuning, making prompt design a central part of modern model use and evaluation \cite{b1,b2,b3,b4}. Among many prompting strategies, persona or role prompting is especially popular because it is intuitive: users can ask a model to respond like a doctor, lawyer, psychologist, scientist, or teacher and expect the answer to reflect domain expertise \cite{b11,b12,b13,b14}. Variants of role prompting are already embedded in many deployed systems through system prompts and assistant framing.

Despite its popularity, the practical value of persona prompting remains unclear. Existing work reports mixed conclusions. Some studies suggest that role prompting can improve reasoning, structure, or domain adaptation \cite{b11,b12,b13}, while others find little or no consistent improvement from persona-based system prompts on factual tasks \cite{b14}. This inconsistency suggests that the practical question is not simply whether persona prompting works, but what kinds of responses it changes, when those changes are beneficial, and how such effects should be evaluated.

A central difficulty is that open-ended answer quality is inherently multidimensional. Recent evaluation work argues that single aggregate scores can obscure important tradeoffs across different response qualities and skills \cite{b21,b22,b26}. This issue is especially relevant for persona prompting. Expert-role prompts may encourage deeper domain framing, more cautious guidance, and more structured reasoning, while simultaneously introducing additional hedging, verbosity, or jargon that reduce readability and directness. If these movements offset each other, aggregate averages may incorrectly suggest that persona prompting has little effect \cite{hu2026expert}.

This perspective also changes how retrieval should be viewed. If role prompting only helps conditionally, then selecting the appropriate expert role becomes more important than simply adding a persona prompt. Most retrieval-augmented generation literature focuses on retrieving factual documents, evidence passages, or external tools \cite{b17,b18,b19,b20}. Our setting differs in that we retrieve expert-role prompts rather than factual knowledge. This reframes retrieval as a role-selection problem: whether selecting a better expert frame changes the quality tradeoffs induced by persona prompting.

To study these questions, we evaluate four prompting and retrieval conditions across 1,140 open-ended questions spanning 38 expert roles and six domains. We compare a baseline without role prompting, a generic domain-expert prompt, embedding-only role retrieval, and a hybrid retrieval approach that combines embedding search with LLM-based role selection. Our main finding is that aggregate averages substantially understate what persona prompting actually does. While overall score differences are small, metric-level analysis reveals a consistent expertise-depth versus clarity tradeoff across conditions.

The effects are strongly conditional. Role prompting performs best on advisory-style interactions and in domains such as medicine and psychology \cite{b35,b37}, where users benefit from cautious expert framing and structured guidance. In contrast, baseline prompting performs better on conceptual and explanatory questions in finance, legal, science, and technology domains, where concise plain-language explanation is often more valuable than specialist framing. We further show that hybrid retrieval improves over embedding-only role selection, although better retrieval does not eliminate the broader tradeoff.

This paper makes three contributions. First, it provides a large-scale controlled comparison of persona prompting and retrieval strategies across 38 expert roles, six domains, and 1,140 open-ended questions. Second, it demonstrates that multi-metric evaluation is necessary because aggregate averages obscure systematic tradeoffs between expertise depth and clarity. Third, it shows that improved role selection reduces some weaknesses of embedding-only retrieval, while also illustrating that the benefits of persona prompting remain conditional rather than universal.

\section{Related Work}
\subsection{Persona and Role Prompting}
Prompting has become a dominant interface for steering large language models at inference time \cite{b1,b2,b3}. Prior work has shown that prompting strategies can substantially alter model behavior and response style \cite{b17,b18,b19,b20}. Within this broader landscape, persona and role prompting provide an intuitive mechanism for inducing domain-specific behavior without model fine-tuning \cite{luz2025persona}.

Prior findings on persona prompting remain mixed. RoleLLM and related work suggest that role prompting can improve reasoning, domain adaptation, and role-consistent behavior \cite{b11,b12,b13}. In contrast, Zheng et al. find that personas embedded in system prompts do not reliably improve factual question-answering performance \cite{b14}.Recent work on sociodemographic persona prompting similarly reports that outcomes are highly sensitive to prompt formulation and evaluation design \cite{b15}.

Our work differs from prior persona-prompting studies in three ways. First, we focus specifically on expert-role prompting rather than fictional or demographic personas. Second, we compare multiple role-selection strategies, including retrieval-based methods. Third, we evaluate open-ended responses through multiple quality dimensions rather than relying primarily on aggregate accuracy or preference scores.

\subsection{Retrieval-Augmented Prompt Selection}
Most retrieval-augmented generation (RAG) research focuses on retrieving factual documents or external evidence to supplement parametric model knowledge \cite{b17,b18,b19,b20}. Our setting differs because we retrieve expert-role prompts rather than factual knowledge. This reframes retrieval as a prompt-selection problem rather than a document-retrieval problem.

Our hybrid retrieval design combines embedding retrieval with LLM-based role selection. The goal is not to compete with conventional RAG systems, but to study whether improved expert-role selection changes the behavioral tradeoffs induced by persona prompting.

\subsection{Multi-Metric Evaluation and LLM-as-a-Judge}
Evaluation of LLM systems is increasingly recognized as a multidimensional problem rather than a single-number ranking problem \cite{b21,b22}. HELM and FLASK argue that aggregate evaluation can obscure important differences across response qualities and capabilities \cite{b21,b22, b36}. This perspective strongly motivates our use of multiple evaluation dimensions rather than relying solely on aggregate preference scores.

The growth of open-ended generation tasks has also accelerated interest in LLM-as-a-judge methods \cite{b23,b24}. At the same time, recent work highlights systematic biases in automated evaluation, including verbosity effects that can artificially reward longer responses \cite{b26}.

Our evaluation framework builds directly on this literature. Rather than collapsing performance into a single preference score, we separately evaluate accuracy, expertise depth, relevance, safety, clarity, and time-sensitive correctness. Our central methodological claim is that metric decomposition makes the behavioral effects of persona prompting substantially more visible.

\section{Method}
\subsection{Experimental Overview}
The goal of this study is to isolate how expert-role prompting changes open-ended response quality under controlled conditions. We compare four prompting strategies that vary in how expert framing is introduced: no role prompt, a generic domain-expert prompt, embedding-only role retrieval, and hybrid retrieval with LLM-based role selection.

Each question is answered under all four conditions using the same generation model. Responses are then evaluated using blind LLM-based judging across six quality dimensions: accuracy, expertise depth, relevance, safety, clarity, and time-sensitive correctness.

\begin{figure}[htbp]
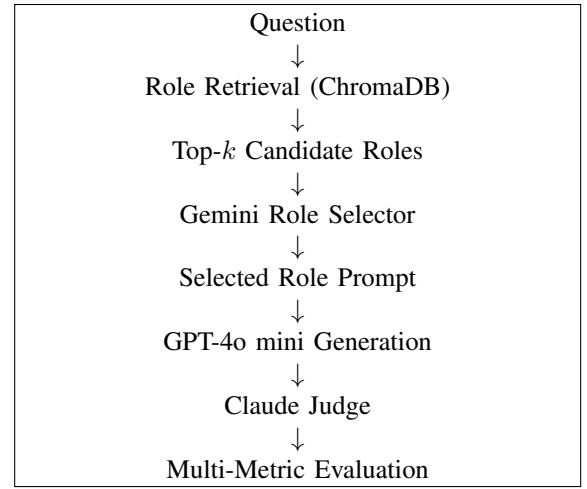

\centerline{%
\fbox{%
\begin{minipage}{0.82\columnwidth}
\centering
Question\\
$\downarrow$\\
Role Retrieval (ChromaDB)\\
$\downarrow$\\
Top-$k$ Candidate Roles\\
$\downarrow$\\
Gemini Role Selector\\
$\downarrow$\\
Selected Role Prompt\\
$\downarrow$\\
GPT-4o mini Generation\\
$\downarrow$\\
Claude Judge\\
$\downarrow$\\
Multi-Metric Evaluation
\end{minipage}}}
\caption{Overview of the hybrid role-retrieval and evaluation pipeline. Baseline and general-expert conditions bypass retrieval and directly prompt the generation model.}
\label{fig:system_overview}
\end{figure}

Figure~\ref{fig:system_overview} illustrates the overall prompting and evaluation pipeline.

\subsection{Role Prompt Corpus}
We construct a curated corpus of 38 expert-role prompts spanning six domains: medical, psychology, finance, legal, science, and technology. Each role is represented as a reusable expert-style system prompt intended to induce domain-specific framing and communication patterns.

Example roles include cardiologist, addiction counselor, clinical psychologist, economist, corporate lawyer, physicist, software architect, and cybersecurity analyst. The prompts are intentionally designed to encourage specialist framing, professional terminology, and domain-appropriate communication styles while avoiding explicit chain-of-thought instructions or task-specific demonstrations.

\subsection{Prompting Conditions}
\textbf{Baseline.} The baseline condition uses no explicit role prompt. Questions are presented directly to the answering model using a neutral system instruction.

\textbf{General expert prompting.} This condition prepends a generic domain-level expert instruction, such as ``You are an expert economist'' or ``You are an experienced cardiologist,'' without retrieval or specialization beyond the broad domain category.

\textbf{Embedding-only retrieval.} In the embedding-only condition, questions are embedded using OpenAI text embeddings and matched against the role-prompt corpus using nearest-neighbor retrieval in ChromaDB. The top retrieved role prompt is injected into the system prompt before answer generation.

\textbf{Hybrid retrieval.} The hybrid condition augments embedding retrieval with an LLM-based selector. First, the top-$k$ candidate roles are retrieved through embedding similarity. Gemini Flash then selects the most appropriate role from the shortlist based on semantic alignment with the user question. The selected role prompt is subsequently injected into generation.

\subsection{Dataset Construction}
The benchmark consists of 1,140 open-ended questions covering 38 expert roles across six domains. Questions are intentionally designed to span both advisory and conceptual interaction types.

Advisory questions focus on guidance, uncertainty management, decision-making, and risk communication. Examples include medical symptom interpretation, psychological coping strategies, financial planning concerns, and legal risk assessment. Conceptual questions focus on explanation, mechanism understanding, and educational clarification.

Representative examples are shown in Table~\ref{tab:dataset_examples}.

\begin{table}[htbp]
\caption{Representative benchmark questions across domains and interaction types.}
\label{tab:dataset_examples}
\begin{center}
\tiny
\begin{tabular}{|p{0.95cm}|p{1.35cm}|p{1.0cm}|p{3.45cm}|}
\hline
\textbf{Domain} & \textbf{Role} & \textbf{Type} & \textbf{Sample Question} \\
\hline
Medical & Cardiologist & Advisory & ``I've been having chest pain for three days that gets worse when I climb stairs. Should I be worried?'' \\
\hline
Finance & Economist & Conceptual & ``How does raising interest rates actually bring down inflation?'' \\
\hline
Psychology & Clinical Psychologist & Advisory & ``I've been feeling emotionally numb for months. Could this be depression?'' \\
\hline
Technology & Software Architect & Conceptual & ``What is the difference between monolithic and microservice architectures?'' \\
\hline
\end{tabular}
\end{center}
\end{table}

The benchmark is intentionally role-structured to provide controlled coverage across domains and interaction types while approximating realistic assistant interactions.

\subsection{Models and Evaluation}
All responses are generated using GPT-4o mini to maintain a consistent answering model across conditions. Hybrid role selection uses Gemini Flash, separating retrieval-time role disambiguation from response generation.

Responses are evaluated through blind LLM-based judging using Claude Haiku 4.5 across six dimensions: accuracy, expertise depth, relevance, safety, clarity, and time-sensitive correctness. Each metric is scored independently on a five-point scale. Judge prompts are structured to encourage metric-specific evaluation rather than overall preference ranking.

This metric decomposition is central to the study because aggregate averages alone can obscure systematic tradeoffs between response qualities. In particular, persona prompting may simultaneously improve expertise depth while reducing clarity or conciseness, causing aggregate scores to understate meaningful behavioral changes.

To reduce positional bias, responses are anonymized and randomized before judging. The evaluation pipeline records both numerical scores and judge justifications for subsequent analysis.

\subsection{Statistical Analysis}
We use nonparametric statistical methods because the score distributions are ordinal and paired across prompting conditions.

Overall differences across conditions are evaluated using the Friedman test. Pairwise comparisons use Wilcoxon signed-rank tests with Holm correction for multiple comparisons. Domain-specific and question-type analyses additionally report effect sizes using Cohen's \emph{d} where appropriate.

We further analyze response length, retrieval similarity, and role-selection behavior to better understand the mechanisms underlying observed metric tradeoffs. Spearman correlation analysis is used for retrieval-quality studies because the relationship between similarity scores and downstream response quality is not assumed to be linear.

\section{Results}
\subsection{Aggregate Results Show Only Small Net Differences}
Table~\ref{tab:overall} summarizes aggregate performance across the four prompting conditions. Baseline achieves the highest average score (4.390), followed closely by hybrid retrieval (4.382), general expert prompting (4.373), and embedding-only retrieval (4.349). A Friedman test detects a statistically significant difference across conditions ($\chi^2 = 31.30$, $p = 7.34 \times 10^{-7}$), although all aggregate effect sizes remain small (all Cohen's \emph{d} $< 0.12$).

Pairwise Wilcoxon signed-rank tests show that baseline significantly outperforms embedding-only retrieval ($p_{adj}=0.0056$), while hybrid retrieval significantly outperforms embedding-only retrieval ($p_{adj}=0.0048$). No significant difference is observed between baseline and hybrid retrieval ($p_{adj}=1.0$).

At the aggregate level, these differences appear modest. However, aggregate averages substantially obscure how persona prompting redistributes performance across individual quality dimensions.

\begin{table}[htbp]
\caption{Overall judged scores across the four conditions.}
\label{tab:overall}
\begin{center}
\scriptsize
\begin{tabular}{|l|c|c|c|c|c|c|c|}
\hline
\textbf{Condition} & \textbf{Acc.} & \textbf{Depth} & \textbf{Rel.} & \textbf{Safe.} & \textbf{Clar.} & \textbf{Time} & \textbf{Avg.} \\
\hline
Baseline & \textbf{4.054} & 3.638 & 4.749 & 4.693 & \textbf{4.896} & 3.952 & \textbf{4.390} \\
Embedding & 3.986 & 3.896 & 4.725 & 4.780 & 4.514 & 3.922 & 4.349 \\
General expert & 4.052 & 3.823 & 4.728 & 4.626 & 4.716 & \textbf{3.957} & 4.373 \\
Hybrid & 4.005 & \textbf{3.923} & \textbf{4.780} & \textbf{4.809} & 4.550 & 3.911 & 4.382 \\
\hline
\end{tabular}
\end{center}
\end{table}

\subsection{Metric Decomposition Reveals the Central Tradeoff}
The clearest pattern emerges after decomposing aggregate scores into individual evaluation dimensions. Role prompting consistently improves expertise depth relative to baseline. Baseline averages 3.638 on expertise depth, compared with 3.896 for embedding-only retrieval, 3.823 for general expert prompting, and 3.923 for hybrid retrieval. The largest improvement is hybrid over baseline, corresponding to a difference of 0.285 (Cohen's \emph{d}=0.54, $p<0.001$).

Clarity moves in the opposite direction. Baseline substantially outperforms all role-injected conditions on readability and directness, scoring 4.896 compared with 4.514 for embedding-only retrieval, 4.716 for general expert prompting, and 4.550 for hybrid retrieval ($p<0.001$ across comparisons).

Accuracy shows a smaller but consistent preference toward the less specialized conditions. Baseline and general expert prompting both score approximately 4.05, while embedding-only and hybrid retrieval remain closer to 4.00. Safety is the one dimension where hybrid retrieval performs best overall, reaching 4.809.

Taken together, these findings suggest that persona prompting does not produce a uniform quality improvement. Instead, it redistributes performance across dimensions. Role prompting tends to generate responses that are more professionally framed, more exhaustive, and more terminology-rich, while simultaneously reducing readability and conciseness. The overall net effect appears modest largely because expertise-depth gains and clarity losses partially offset each other in aggregate averages.

\begin{figure}[htbp]
\centerline{\includegraphics[width=0.95\columnwidth]{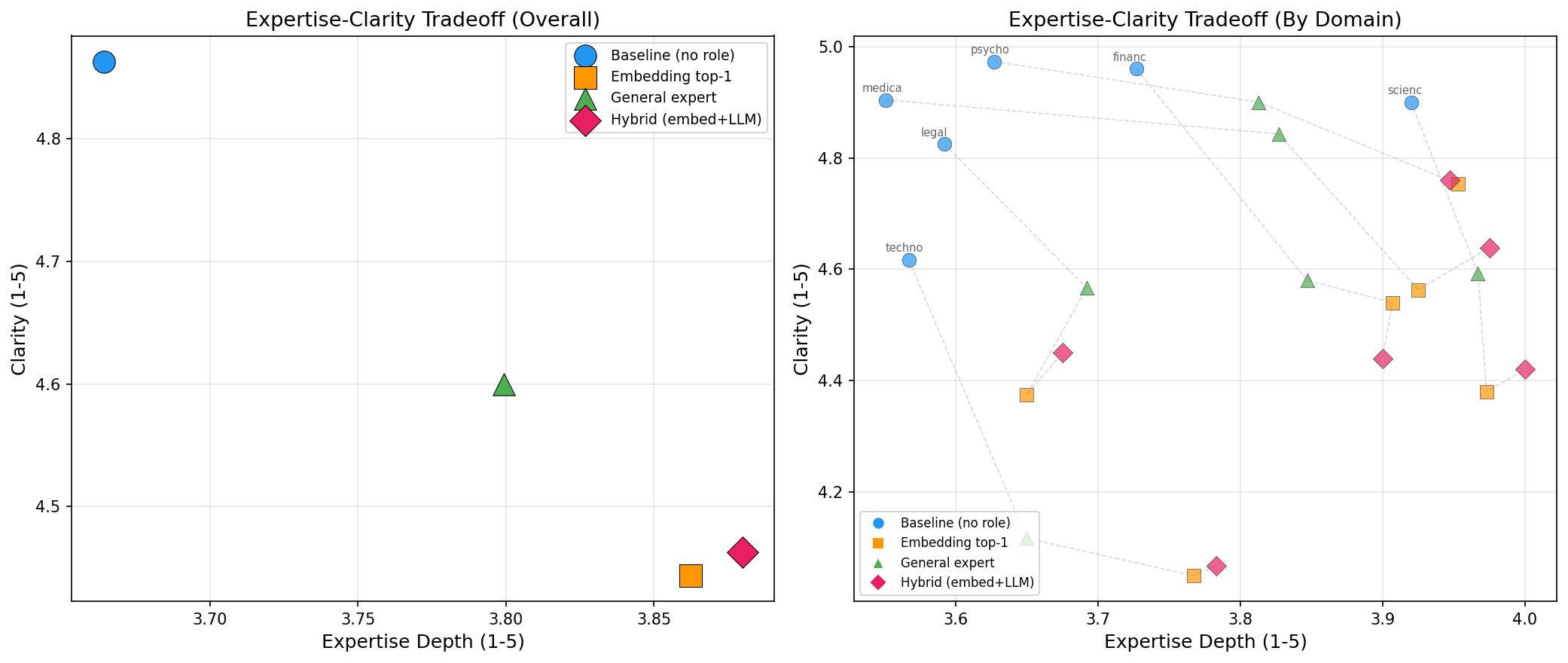}}
\caption{Expertise-depth versus clarity tradeoff across prompting conditions. Role-injected conditions consistently move toward greater expertise depth at the cost of reduced clarity. The effect is strongly domain-dependent: medical and psychology settings benefit more from expert framing, whereas finance, legal, science, and technology domains remain closer to the high-clarity baseline region.}
\label{fig:pareto}
\end{figure}

Figure~\ref{fig:pareto} visualizes this expertise-depth versus clarity tradeoff. Role-injected conditions consistently shift toward greater expertise depth at the cost of reduced clarity. The effect is strongly domain-dependent: medical and psychology settings benefit more from expert framing, whereas finance, legal, science, and technology domains remain closer to the high-clarity baseline region.

\subsection{Question Type Explains When Role Prompting Helps}
Question type provides the clearest explanation for the conditional nature of persona prompting. On advisory questions ($n=850$), hybrid retrieval performs best with an average score of 4.403, followed by general expert prompting (4.387), baseline (4.376), and embedding-only retrieval (4.367).

Advisory questions typically involve uncertainty management, risk assessment, or guidance-oriented reasoning. In these settings, expert framing appears to encourage more structured recommendations, cautious communication, and professional reasoning patterns. These characteristics often improve perceived expertise depth and safety, particularly in medical and psychological contexts.

The ordering reverses on conceptual questions ($n=284$). Baseline performs best at 4.435, ahead of general expert prompting (4.330), hybrid retrieval (4.319), and embedding-only retrieval (4.297). Conceptual questions reward concise explanation and conceptual clarity more than specialist framing. Under these conditions, role prompting often introduces additional terminology, hedging, or unnecessary detail that increases perceived expertise while reducing readability.

This contrast suggests that the usefulness of persona prompting depends less on the domain label itself and more on whether the interaction benefits from expert framing as part of the task.

\subsection{Domain Effects Are Strongly Asymmetric}
The domain breakdown further reinforces the conditional nature of persona prompting. Hybrid retrieval achieves the highest average score in medical (4.457) and psychology (4.445), compared with baseline scores of 4.394 and 4.389 respectively. The medical improvement is statistically strong, while psychology shows a smaller but directionally consistent improvement.

In contrast, baseline performs best in finance (4.364), legal (4.254), science (4.541), and technology (4.323). Across all six domains, baseline also remains the clearest condition, while role-injected variants consistently increase expertise depth.

One possible explanation is that medicine and psychology are inherently advisory domains where users value structured caution, uncertainty communication, and professional framing. Finance, science, legal, and technology questions in the benchmark are comparatively more conceptual and explanatory, rewarding concise plain-language communication instead.

These findings are consistent with the interpretation that persona prompting helps most when specialist framing itself contributes meaningful value to the interaction.

\subsection{Hybrid Retrieval Consistently Improves Over Embedding-Only Retrieval}
Across nearly all analyses, hybrid retrieval outperforms embedding-only retrieval. Aggregate scores improve from 4.349 to 4.382, while expertise depth increases from 3.896 to 3.923. Hybrid retrieval also improves safety and relevance relative to embedding-only retrieval.

These results suggest that retrieval quality matters. Embedding similarity alone often retrieves semantically adjacent but imperfectly aligned expert roles. Introducing an LLM-based selector over the retrieved shortlist appears to improve role disambiguation and contextual alignment.

At the same time, improved role selection does not eliminate the broader expertise-depth versus clarity tradeoff. Hybrid retrieval remains substantially less clear than baseline prompting despite its stronger depth and safety performance. This pattern suggests that the tradeoff is not solely a retrieval-quality artifact, but may instead reflect a broader behavioral consequence of expert-role prompting itself.

\subsection{Response Length and Verbosity Effects}
Response length analysis reveals a consistent verbosity increase under role prompting. Hybrid and embedding-only retrieval conditions generate substantially longer responses on average than baseline prompting.

Qualitative inspection suggests that these responses frequently include additional caveats, specialist terminology, structured recommendations, and broader contextual framing. In advisory settings, these characteristics often contribute positively to perceived expertise and safety. In conceptual settings, however, they frequently reduce readability and directness without substantially improving informational content.

This observation connects the quantitative metric tradeoff to a concrete linguistic mechanism. The expertise-depth gains associated with persona prompting appear closely tied to changes in discourse style rather than purely factual improvements.

\section{Discussion}
Our results suggest that persona prompting changes LLM behavior more reliably than it improves overall capability. Across nearly all analyses, expert-role prompting systematically increases expertise depth while reducing clarity and directness. This tradeoff becomes difficult to detect when evaluation is collapsed into a single aggregate score, since expertise gains and clarity losses partially offset one another.

One interpretation is that persona prompting changes not only what information the model provides, but also how the model frames expertise itself. Expert-role prompts encourage more cautious language, broader contextual framing, and more structured recommendations. These behaviors often improve perceived professionalism and depth in advisory settings such as medicine and psychology, where users value uncertainty management and risk communication. In contrast, conceptual and explanatory questions in finance, legal, science, and technology domains tend to reward concise communication and readability more than specialist framing. Under these conditions, role prompting frequently introduces additional terminology and verbosity without proportionally improving informational content.

Our findings also reinforce the importance of multidimensional evaluation for open-ended generation systems. Aggregate preference scores alone substantially understate the behavioral effects of persona prompting. A system may appear nearly unchanged under aggregate evaluation while simultaneously becoming more specialized, more verbose, safer, or less readable. More broadly, these results suggest that persona prompting should not necessarily be treated as a universally beneficial default strategy. Instead, its usefulness appears highly dependent on interaction type and user objective.

\section{Limitations}
This study has several limitations. First, the benchmark is synthetic and role-structured rather than drawn from naturally occurring user traffic. Although the questions were designed to reflect realistic interaction patterns, real-world conversational distributions may differ substantially.

Second, the study evaluates a limited set of models and prompting configurations. Different models may exhibit different sensitivities to role prompting, retrieval quality, or verbosity effects, particularly smaller or open-source systems.

Third, although blind LLM-based judging enables scalable evaluation, automated evaluation remains imperfect. Judge models may exhibit latent preferences for verbosity, structure, or professional terminology despite response anonymization and metric decomposition. The results should therefore be interpreted as controlled comparative evidence rather than definitive human preference measurements.

\section{Conclusion}
This paper studies when persona prompting actually improves open-ended LLM responses. Through a controlled comparison of four prompting conditions across 1,140 questions, 38 expert roles, and six domains, we find that persona prompting produces only small aggregate differences but substantial multidimensional behavioral changes.

The dominant pattern is a consistent expertise-depth versus clarity tradeoff. Role prompting systematically increases perceived expertise, structure, and professional framing while reducing readability and directness. These effects are highly conditional rather than universal. Persona prompting performs best in advisory settings such as medicine and psychology, where structured expert framing contributes meaningful value to the interaction. In contrast, baseline prompting often performs better for conceptual and explanatory questions that prioritize concise communication.

We further show that hybrid retrieval improves over embedding-only role selection, although better retrieval does not eliminate the broader tradeoff induced by expert-role prompting. More generally, our results demonstrate that aggregate evaluation alone can obscure important behavioral effects in open-ended generation systems.

Overall, our findings suggest that persona prompting primarily reshapes response characteristics rather than broadly improving capability. Understanding these tradeoffs requires multidimensional evaluation frameworks capable of capturing how prompting changes not only what models say, but how they communicate expertise itself.

\section*{Acknowledgment}
The authors thank the project collaborators whose scripts, evaluation artifacts, and analysis outputs made this draft possible.

\bibliographystyle{IEEEtran}
\bibliography{reference}

\end{document}